\documentclass{article}

\usepackage[letterpaper]{geometry}
\usepackage{microtype}
\usepackage{amsmath}
\usepackage{graphicx}
\usepackage{color}
\usepackage{xspace}
\usepackage[colorlinks]{hyperref}
\usepackage[font=small]{caption}

\makeatletter
\DeclareRobustCommand\onedot{\futurelet\@let@token\@onedot}
\def\@onedot{\ifx\@let@token.\else.\null\fi\xspace}

\def\eg{\emph{e.g}\onedot} 
\def\ie{\emph{i.e}\onedot}

\def\etal{\emph{et al}\onedot}
\makeatother

\title{Fully-Automatic Synapse Prediction and Validation on a Large Data Set}

\author{Gary B. Huang, Louis K. Scheffer, Stephen M. Plaza}

\date{\today}

\begin{document}
\maketitle

\begin{abstract}
Extracting a connectome from an electron microscopy (EM) data set requires identification of neurons and determination of connections (synapses) between neurons.  As manual extraction of this information is very time-consuming, there has been extensive research effort to automatically segment the neurons to help guide and eventually replace manual tracing.  Until recently, there has been comparatively less research on automatically detecting the actual synapses between neurons.  This discrepancy can, in part, be attributed to several factors: obtaining neuronal shapes is a prerequisite first step in extracting a connectome, manual tracing is much more time-consuming than annotating synapses, and neuronal contact area can be used as a proxy for synapses in determining connections.

However, recent research has demonstrated that contact area alone is not a sufficient predictor of synaptic connection \cite{Takemura2013}.  Moreover, as segmentation has improved, we have observed that synapse annotation is consuming a more significant fraction of overall reconstruction time (upwards of 50\% of total effort).  This ratio will only get worse as segmentation improves, gating overall possible speed-up.  Therefore, we address this problem by developing algorithms that automatically detect pre-synaptic neurons and their post-synaptic partners.  In particular, pre-synaptic structures are detected using a Deep and Wide Multiscale Recursive Network (DAWMR), and post-synaptic partners are detected using a Multilayer Perceptron (MLP) with features conditioned on the local segmentation.

This work is novel because it requires minimal amount of training, leverages advances in image segmentation directly, and provides a complete solution for polyadic synapse detection.  We further introduce novel metrics to evaluate our algorithm on connectomes of meaningful size.  These metrics demonstrate that complete automatic prediction can be used to effectively characterize most connectivity correctly.
\end{abstract}

\section{Introduction}
High-resolution, EM imaging allows one to identify synapses, such as the one shown in Figure \ref{fig:synapse_ex} below.  In this example, there is a electron dense region corresponding to the synapse at the pre-synaptic body.  This consists of different transport apparatus, such as vesicles, that abut the neuronal membrane.  In a data set that contains numerous organelles of varying electron density (\ie, imaging intensity) and neuronal membrane that intricately weaves throughout, identifying synapses can be challenging.  When creating a connectome, an annotator will typically scan the data set or a traced neuron and manually identify and mark these sites.  Even for organisms as small as a fruit fly, there are upwards of 100 million connections, making the process of manual annotation intractable.

\begin{figure}
\centering
\includegraphics[width=0.5\textwidth]{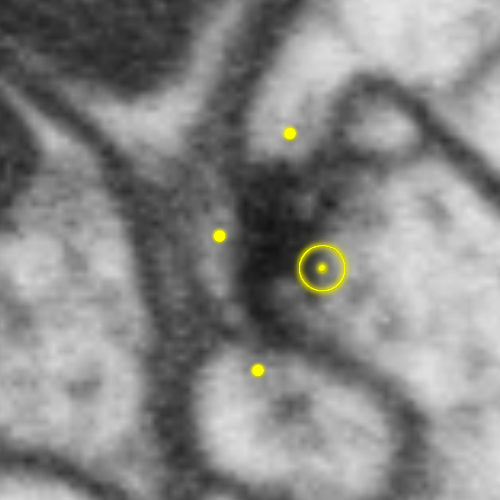}
\caption{\label{fig:synapse_ex} An example of a synapse in the Drosophila optic lobe.  The larger yellow bullseye marks the pre-synaptic structure, which we refer to as a T-bar due to its shape, and the smaller yellow dots mark three post-synaptic densities (PSDs) that partner with the marked T-bar.}
\end{figure}

Consequently, there have been recent research efforts to automate synapse detection using machine learning, which we discuss below in Section~\ref{sec:background}.  However, existing techniques for automated synapse detection have primarily been applied to detection in mammalian tissue.  It is unclear, then, how well such approaches would translate when applied to synapse detection in Drosophila tissue.  In contrast to synapses in the mammalian brain, which are predominantly monadic, involving a single pre-synaptic site and post-synaptic site, synapses in the Drosophila are mostly polyadic, involving multiple post-synaptic partners for a given pre-synaptic site~\cite{Cardona2010}.  For example, Figure~\ref{fig:synapse_ex} shows a pre-synaptic site with three post-synaptic partners.  These neuronal processes are often difficult to segment, which makes identifying the post-synaptic partners non-trivial, even given the pre-synaptic site.


Prior work on automated synapse identification has also focused only on the detection problem in isolation.  However, synapse detection is one step in a larger pipeline, whose final goal is the extraction of a connectome from the EM data.  Therefore, we are interested not only in individual synapse detection accuracy, but in how synapse detection integrates into this larger system, and how errors in individual steps in that system combine when evaluating the final produced connectome.

For instance, one straightforward method for reliably using automated synapse detections in an EM pipeline is as hints for manual annotation, as done by Plaza~\etal~\cite{Plaza2014}.  By manually verifying detections, errors in the final connectome are minimized, but at the expense of human effort and time.  An alternative would be to simply accept all detected synapses above a certain confidence threshold, but there has been no prior work on whether such a prediction would result in a meaningful connectome.  In particular, many connections between neurons are formed from a high number of synaptic contacts, and therefore one might hope that automated algorithms are capable of faithfully reconstructing such high strength connections, but this has not yet been experimentally tested.

Moreover, extracting a connectome is also dependent upon automatic neuron segmentation.  In addition to possibly being outright incorrect, a segmentation may also be noisy along a border.  Both cases may potentially cause errors in the connectivity graph when combined with automated synapse identification output.  


Therefore, in this paper, we introduce algorithms that enable fully automatic synapse prediction and evaluate the results of the end-to-end process from the standpoint of the final produced connectome.  Specifically, key contributions of our approach and results include:

\begin{enumerate}
\item an algorithm that generalizes well over a large data set with minimal supervision required,
\item new metrics to better evaluate synapse prediction in realistic settings,
\item empirical results analyzing end-to-end accuracy of the proposed approach on a publicly available connectome data set~\cite{Takemura2015}, demonstrating high performance and preservation of biological pathways, in particular relative to a baseline using body-proximity as a proxy for synaptic contact.
\end{enumerate}

\section{Background}\label{sec:background}

An automated approach for synapse identification in EM images using machine learning was first proposed by Kreshuk~\etal, who use a Random Forest (RF) classifier on hand-selected image features to detect synapses~\cite{Kreshuk2011}.  In subsequent work, Kreshuk~\etal extend this method by applying graph cut on the synapse probabilities to obtain a segmentation of each putative synapse, extracting object-level features given the segmentation, and then applying a RF classifier to determine whether each segmented region is a synapse or not~\cite{Kreshuk2014}.

Becker~\etal attempt to generate more informative features, by conditioning on the synaptic cleft, thereby allowing features to be extracted from consistent spatial locations relative to the putative synapse~\cite{Becker2013}.  These features are then used as input to AdaBoost for synapse detection.  

Jagadeesh~\etal consider the problem of large-scale synapse detection in a large image volume~\cite{Jagadeesh2013}.  They first use a fast interest point detector based on image-thresholding to generate proposals for possible synapse locations.  They then use feature descriptors hand-designed to extract information about relevant biological biological structures, namely vesicles, clefts, and ribbons.  These features are used as input to an SVM or Multiple Kernel Learner for patch-based synapse detection.

Biological preparation has also been considered as a means toward aiding in automated synapse detection.  Navlakha~\etal apply a technique for selectively staining for synapses, leading to more pronounced opacity at synaptic sites, and leaving non-synaptic membranes unstained~\cite{Navlakha2013}.  They propose a high-throughput method for automated detection by first filtering down to a candidate set of patches, then applying an SVM to classify each patch as synapse or non-synapse. While this technique can be used to compute statistics on synapses such as density, since the membranes are left unstained, it cannot be used in conjunction with segmentation and therefore cannot be directly used for extracting a connectome.

Most recently, Roncal~\etal also consider large-scale synapse detection, presenting two different techniques~\cite{Roncal2015}.  They propose a fast RF classifier using hand-selected features, including a filter designed for vesicle detection.  This RF classifier yields similar results to Becker~\etal~\cite{Becker2013}, but with approximately half the run-time.  They also propose a deep learning classifier for synapse detection, which yields superior results to the fast RF classifier but approximately two orders of magnitude slower.

The above approaches were evaluated on synapse detection in mammalian tissue, assuming a single post-synaptic site for each pre-synaptic site.  Several approaches also make additional assumptions on the data, such as being able to reliably identify the synaptic cleft to extract spatially consistent features~\cite{Becker2013}, or having feature descriptors hand-tuned for particular biological structures~\cite{Jagadeesh2013}.  The method of Kreshuk~\etal~\cite{Kreshuk2011} was adapted for pre-synaptic site detection/proposal by Plaza~\etal~\cite{Plaza2014}, but labeling post-synaptic partners was performed manually.

As mentioned in the introduction, synapse detection in Drosophila can be more challenging, due to the polyadic nature of such synapses, where pre-synaptic sites have multiple post-synaptic partners, and where post-synaptic processes can often be small and difficult to segment.  To address this difficulty in Drosophila synapse detection, Kreshuk~\etal specifically studied the problem of synaptic partner assignment.  Conditioned on ground-truth neuron segmentation and synapse detection, they formulate a pairwise graphical model wherein nodes of the model represent possible assignments between two neurons $i,j$ at a putative synapse, \eg, neuron $i$ is pre-synaptic and neuron $j$ is post-synaptic.  

In this work, we propose a complete system for automated synapse detection, capable of handling polyadic synapses as found in Drosophila.  Our system uses unsupervised feature learning, and takes as training data simple point-wise annotations of pre- and post-synaptic sites, and therefore can be applied to new data sets with relatively minimal supervision.  By comparison, existing methods as discussed above, that require hand-designed features to extract high-level information such as vesicles and ribbons, may not be appropriate for new data sets, or may require significant manual effort to tune or re-design the feature descriptors.  In contrast to Kreshuk~\etal, we evaluate our system on completely-automated, noisy segmentation.

\section{Automated Synapse Detection}

Our system for automated synapse detection proceeds in two distinct steps.  First, independent of any segmentation, we apply a classifier to automatically identify pre-synaptic sites in the Drosophila, which are often referred to as T-bars, due to their T-like shape, formed by a pedestal and platform structure.  Next, conditioned on predicted T-bar locations and a segmentation, we apply a second classifier to predict partnering post-synaptic densities (PSDs) for the identified T-bars.  We describe each of these steps in more detail in the next two sections.  We will also be releasing source code that implements the proposed methods.\footnote{Code will be available at \url{https://bitbucket.org/gbhuang/flyem_synapse}.}

\subsection{Pre-synaptic T-bar Identification Algorithm}\label{sec:tbar}

The first step in our automated synapse detection pipeline is to detect the pre-synaptic T-bar sites.  An example of a T-bar can be seen in Figure~\ref{fig:synapse_ex}.  Due to its distinct structure, we focus on first predicting T-bars in isolation, independent of both segmentation and PSD prediction, and delay the problem of determining the potentially multiple PSD partners until after segmentation, as PSDs are typically more ambiguous and difficult to identify.  We note that this approach of splitting T-bar and PSD prediction into separate steps, with PSD prediction aided by segmentation, has also been taken for manual synapse detection~\cite{Plaza2014}.

For automated T-bar detection, we make use of Deep and Wide Multiscale Recursive (DAWMR) networks~\cite{Huang2014}.  We give the highlights of our approach here; for more details, see Huang and Plaza~\cite{Huang2014a}.

Unlike the problem of image segmentation, which is naturally framed as a voxel-wise prediction problem (at each voxel, predict whether that voxel belongs to a cell boundary or not), T-bar detection is an object detection problem, which we formulate as predicting, for each T-bar, a point annotation, specifying the spatial coordinates of the center of the T-bar.  To generate voxel-wise training data for the DAWMR network, we simply consider any voxels within a certain radius of a T-bar point annotation to be a positive example, and all other voxels to be negative examples.  We find that the DAWMR networks are able to successfully learn from this simple training data, allowing for less manual supervision effort relative to methods and tasks that require dense labeling.

To generate final T-bar point predictions from the voxel-wise output of the DAWMR network, we spatially smooth the voxel-wise predictions, selecting the voxels with highest confidence, and apply non-maxima suppression.

We make two notes concerning evaluation of T-bar prediction, in the context of a larger connectomics pipeline.  First, it is important to consider the precision/recall (PR) curve for the automated predictions.  Different applications may have different misclassification costs, leading to different thresholds along different points of the PR curve.  For instance, if very high fidelity is required, one may need to select a threshold for high recall, at the expensive of precision, whereas if the final goal is to determine strong connections in the connectome with some tolerance for small errors, the optimal threshold may be to select for the precision/recall break even point.

Second, T-bar prediction accuracy can be computed by requiring, for instance, that predicted T-bars be within a specified distance of a ground-truth T-bar to be counted as a correct match, as in Huang and Plaza~\cite{Huang2014a}.  However, ultimately, the exact location of a T-bar annotation will be abstracted as one endpoint of an edge in a connectomic graph, indicating the pre-synaptic body.  Therefore, the primary concern is that annotation be placed in the correct neuron.  Thus, when a segmentation is available, T-bar prediction accuracy should be computed by further requiring that the predicted T-bar fall within the same segment as the ground-truth T-bar.

Due to this interaction with the segmentation when evaluating T-bar performance accuracy, it may be beneficial to post-process the T-bar predictions.  For instance, we find that our T-bar detection often places the annotation in the distinctive dark T-like structure itself, which, due to its dark intensity, can cause problems for automated segmentation.  We therefore find a benefit in slightly shifting T-bar predictions within a small radius to the brightest intensity voxel, helping the annotation be placed in a non-ambiguous region relative to the segmentation.

\subsection{Segmentation-aware Post-synaptic Partner Identification}\label{sec:psd}

Once we have automated T-bar predictions and a (possibly automated) segmentation, we condition on this information in order to predict the post-synaptic densities (PSDs) that partner with each T-bar.  For a given T-bar, we can consider all nearby segments as potentially having a partner PSD.  More precisely, we use the set of segments that have non-empty intersection with a sphere of a given radius, centered at a given T-bar, as the candidate set of bodies that may be post-synaptic to the T-bar.  We exclude the segment containing the T-bar itself, and therefore make no attempt at predicting autapses.  Additionally, we do not attempt to identify cases where a single T-bar makes multiple connections to the same post-synaptic body, and thus any such biological multiple connections will at most be predicted as a single synapse.

With this set-up, we have a binary classification problem, where for each T-bar and each candidate segment, we wish to determine if the candidate segment contains an actual PSD and thus forms a synapse with the T-bar.  For classification, we use a Multi-Layer Perceptron (MLP).  To generate the feature representation, we estimate the interface of the synapse between the T-bar segment and candidate segment, by dilating both segments by varying amounts and letting the estimated interface be the intersection.  We then pool a set of simple image features over the interface, computing statistics such as size of the interface and image intensity within the interface (such as number of voxels with intensity lower than some set threshold).  

One important consideration is that PSD prediction performance will depend on both the accuracy of the PSD predictor itself as well as the performance of the algorithm used to generate the segmentation.  Therefore, it may be necessary to tune the PSD predictor with an awareness of the behavior of the segmentation algorithm.  For instance, we found that dark intensity values such as those found at a boundary, as well as at T-bars, would often present difficulties for the segmentation algorithm.  This ambiguity could lead to, for instance, small parts of the T-bar segment being incorrectly assigned to a neighboring segment.  Although such localized errors would not have a large effect on the topology of the segmentation (in terms of Rand error, for example), they can have a large effect on the proposed feature representation and hence PSD classifier.  Therefore, we attempt to make the classifier more robust to such errors by ignoring the segmentation at voxels with such dark intensity values.

\section{Metrics for Evaluation}\label{sec:metrics}

As discussed above, to properly evaluate automated synapse detection performance in the context of a larger pipeline, it is important to consider the full performance curve as the threshold on classifier confidence is varied.  This allows for synapse prediction to be evaluated at the appropriate threshold for varying misclassification costs, which will depend on the final application being considered.  One straightforward metric for evaluating detection at the individual synapse level is to produce a precision/recall (PR)  curve.  Under the view of the connectome as a graph, with directed edges between nodes (representing neurons) defined by synapses, we can consider two variations for computing PR.  First, we can view the connectome as a weighted graph, and compute PR by considering each individual synapse as a ground-truth label that is to be predicted.  Second, we can consider the connectome as an unweighted graph, and compute PR by considering each edge (formed by any number of synapses between a pair of neurons) as a ground-truth label to be predicted.

The above methods for computing PR are two ways of dealing with the finding from connectomic studies that many connections between neurons consist of multiple synapses~\cite{Takemura2013,Takemura2015}.  This multiplicity may be a weight on the synapse strength, or may be a mechanism for robustness.  In either case, a general assumption in many connectomic efforts is that important biological connections will have some multiplicity greater than one.  Therefore, we would like to consider a range of metrics that will better reflect whether a set of automated synapse predictions is actually good enough for use in connectomic studies.

Computing PR with a weighted graph requires that the automated predictions match the ground-truth precisely in terms of strength, without regard to topology.  For example, predicting an edge of strength 7 for a ground-truth edge of strength 9 is equivalent to missing an edge of strength 2 (in terms of recall value), which may be inappropriate if we do not care about precisely determining the multiplicity of strong connections.  Computing PR with an unweighted graph, on the other hand, evaluates the automated predictions solely in terms of unweighted topology.  Therefore, no penalty is incurred for not correctly determining multiplicity, but missing an edge of multiplicity 1 is equivalent to missing an edge of multiplicity 7.  

One simple modification that can be made to the unweighted graph computation is to consider the unweighted graph produced by thresholding the edge weights by some value $t$ (in both the predicted and ground-truth connectomes).  For $t=1$, we have the original unweighted PR; for $t>1$, we focus only on stronger predicted and ground-truth edges, with multiplicity of at least $t$.  We can also examine performance of a given classifier over different sets of curves as we vary this threshold $t$.

\subsection{Asymmetric PR, Connections Added/Missed}

By thresholding the edge weights at some $t>1$ and computing unweighted PR, we focus on the strong edge connections and ignore potentially noisy weak connections.  However, there is still a strong boundary effect, where, for instance, a predicted edge of strength $t-1$ for a corresponding ground-truth edge of strength $t$ is counted as a false negative, the same as if the predicted edge strength had been simply zero.  This harsh decision boundary may also be problematic from the standpoint of potential small errors in the manually annotated ground-truth.  We would like a metric that focuses on identifying clear error cases in the automated predictions.  

We therefore introduce an asymmetric variant of the above thresholded PR curve.  Let the asymmetric $t_1,t_2$ thresholded PR curve (with $t_1 > t_2$) be defined as follows: consider the (weighted) ground-truth connectome graph $g$ and the predicted graph $p$ produced by applying some classifier threshold, and let $g(e)$ be the weight of a given edge $e$ in $g$, and similarly for $p(e)$.  Recall is then computed as 
$$\frac{\sum_e [p(e) \geq t_2 \wedge g(e)\geq t_1]}{\sum_e [g(e)\geq t_1]},$$ 
where the square Iverson brackets evaluate to 1 if the condition inside is true and 0 otherwise.  In other words, the total set of positive ground-truth instances consists of all edges with ground-truth weight greater than $t_1$, but the subset of true positives allows for edges with predicted weight greater than the smaller $t_2$.  Conversely, precision is computed as 
$$\frac{\sum_e [p(e) \geq t_1 \wedge g(e)\geq t_2]}{\sum_e [p(e)\geq t_1]}.$$  
Here the total set of positive predicted instances consists of all edges with predicted weight greater than $t_1$, but the subset of true positives allows for edges with ground-truth weight greater than the smaller $t_2$.  

From the above precision/recall definitions, it can be seen that asymmetric $t_1,t_2$ thresholded PR upper bounds the original symmetric thresholded PR at $t=t_1$.  This more lenient performance measure focuses on the more clear, egregious errors, where there is a strong edge in either the ground-truth or predicted connectome graph but a weak or no edge in the other graph.  We can also report these types of errors directly as connections (falsely) added and connections missed.  Let connections missed be the set of edges $e$ such that $g(e)\geq t_1 \wedge p(e)<t_2$.  The number of connections missed is an unnormalized version of $1 - \mathrm{recall}$.  Let connections added be the set of edges $e$ such that $p(e)\geq t_1 \wedge g(e)<t_2$.  The number of connections added is an unnormalized version of $1 - \mathrm{precision}$.  When plotting number of connections added versus number of connections missed, we normalize these values by the number of edges in the ground-truth connectome after thresholding, \ie, the number of edges $e$ such that $g(e) \geq t_1$, to put curves with different values of $t_1$ on the same scale.

By using asymmetric thresholded PR and connections added/missed, we can focus on strong error cases when comparing sets of predictions and be robust to small amounts of labeling noise.  These error measures also more clearly indicate to what extent strong biological connections are being missed or false introduced through prediction.

\section{Results}

In this section, we present a case study of our proposed synapse detection system on data from the Drosophila optic lobe.  The data set that we use comprises 7 columns of the medulla.  Details of the data can be found in Plaza~\etal~\cite{Plaza2014} and Takemura~\etal~\cite{Takemura2015}, and the raw EM image data, FIB-25, is available online.\footnote{\url{https://www.janelia.org/project-team/flyem/data-and-software-release}}

We give results of the individual steps of our pipeline, full end-to-end results, results using the proposed error metrics focusing on clear error cases, comparison against a surface area contact baseline, and results in the context of preserving biological findings.

\subsection{Performance of T-bar, PSD detectors}\label{sec:psd_perf}

\begin{figure}[htbp]
\centering
\includegraphics[width=0.45\textwidth]{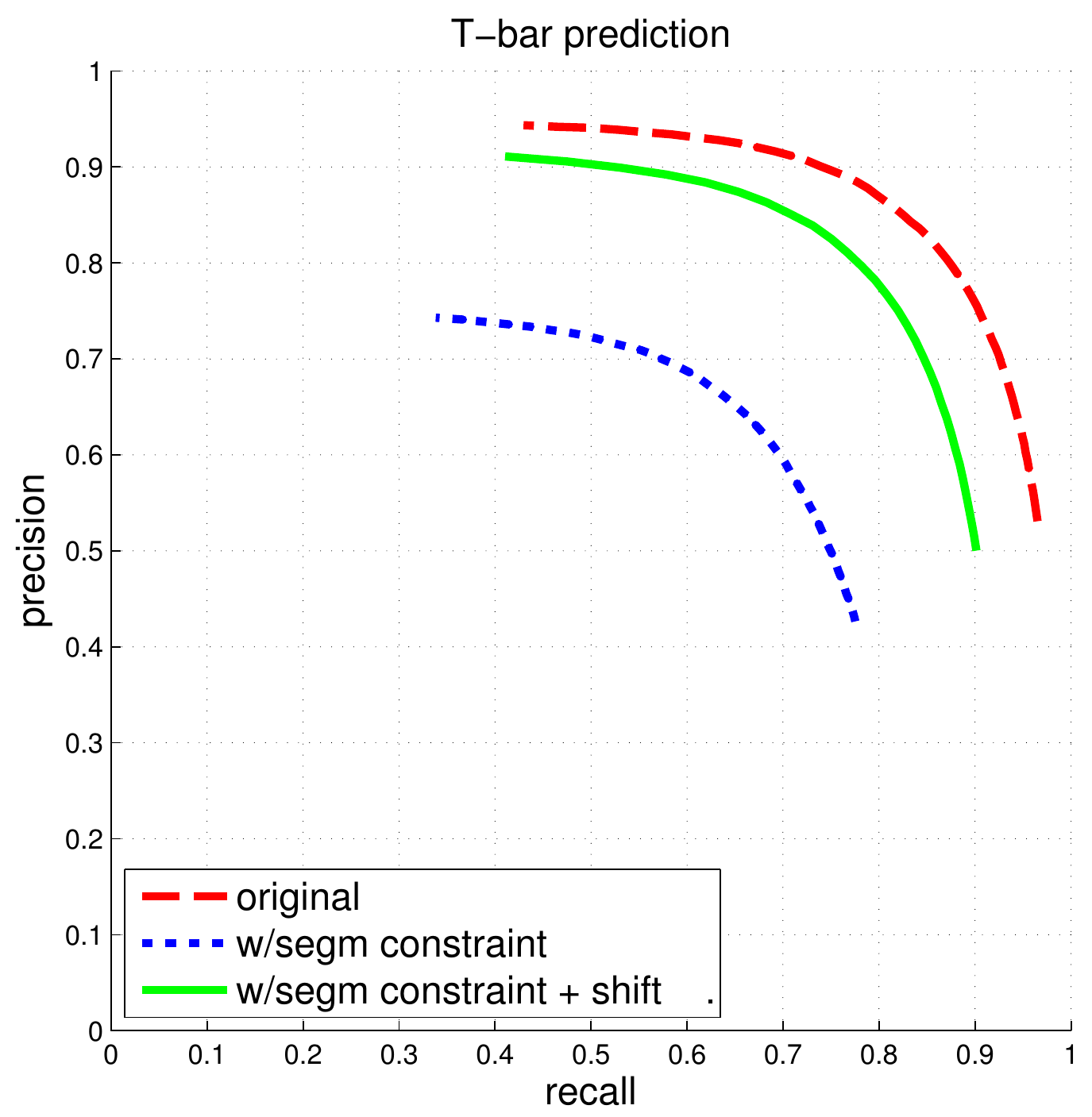}
\caption{\label{fig:fib25_tbar} T-bar precision/recall.  (This figure and subsequent figures best viewed in color.)  The dashed red curve indicates PR when the only constraint for a match between a predicted T-bar and a ground-truth T-bar is that the two locations fall within a specified distance of each other.  The dotted blue curve gives PR when a match is further constrained to enforce that the predicted T-bar and ground-truth T-bar both fall within the same segment, in the ground-truth segmentation.  Finally, the solid green curve gives PR with the segmentation constraint, if the predicted T-bar locations are first shifted slightly, away from potentially ambiguous regions.  See the accompanying text and Section~\ref{sec:tbar} for more discussion.}
\end{figure}

We first train a T-bar detector using the system described above in Section~\ref{sec:tbar}, using the ground-truth annotations contained in one $520^3$ subcube of the total volume.  Figure~\ref{fig:fib25_tbar} gives the precision/recall curve for the automated predictions over the entire data volume.  The plot highlights two important points that were made in Section~\ref{sec:tbar}: First, T-bar prediction accuracy should ideally be assessed within the context of segmentation and the final produced connectome graph, rather than only considering the distance between predicted and ground-truth T-bar locations.  A predicted T-bar that is very close to a ground-truth T-bar, but placed in the wrong ground-truth segment, will lead to errors in the connectome graph.  This is highlighted in the difference between the dashed red curve and the dotted blue curve.  Consequently, second, the T-bar detector may need to be aware of the behavior of the corresponding segmentation algorithm.  In our case, we found that by simply shifting the predicted T-bar locations slightly, toward brighter image voxels, this would move the predictions away from dark image regions that are more difficult or ambiguous for the segmentation algorithm, and therefore improve performance when applying the segmentation constraint.

\begin{figure}
\centering
\includegraphics[width=0.45\textwidth]{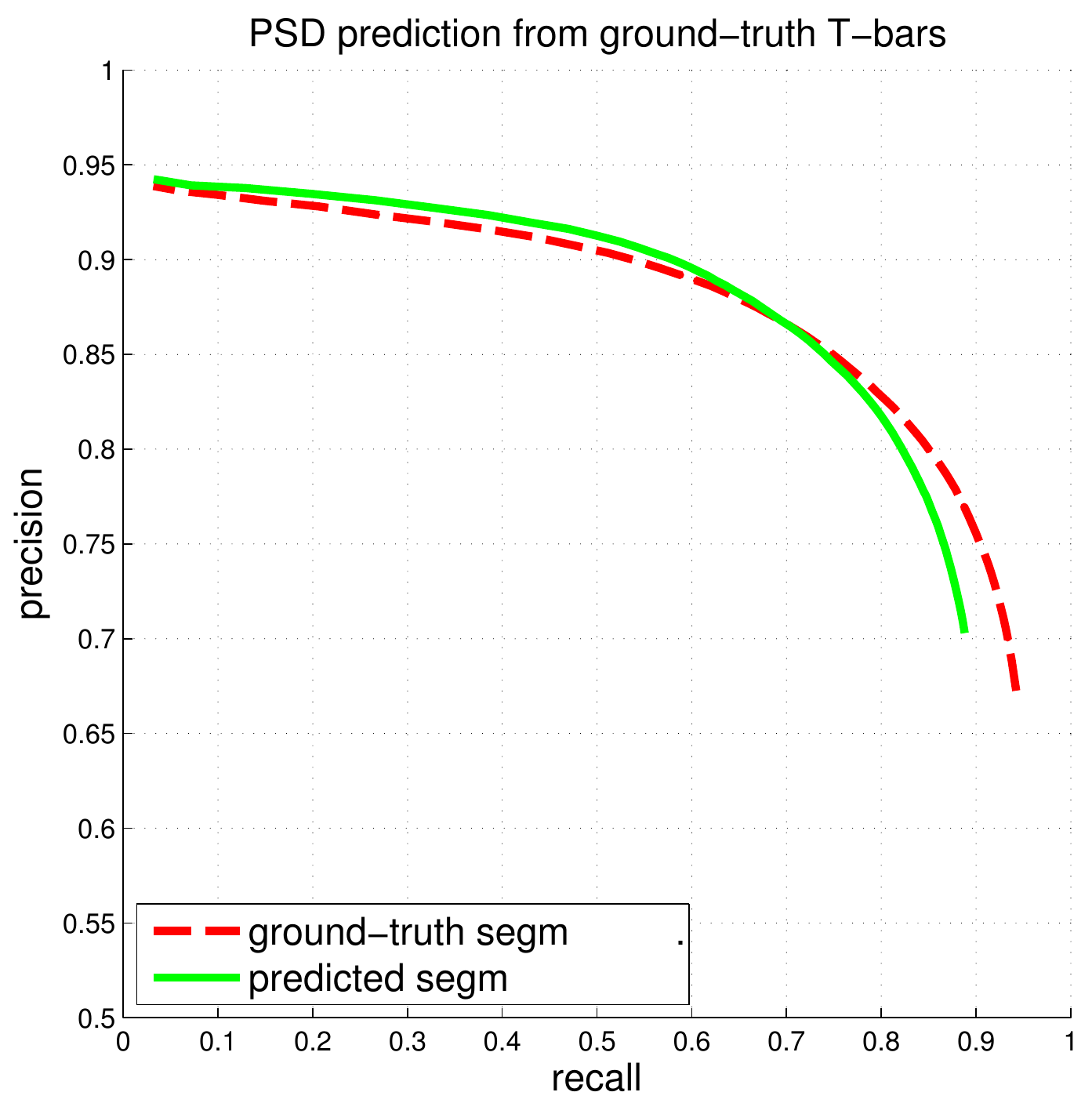}
\includegraphics[width=0.45\textwidth]{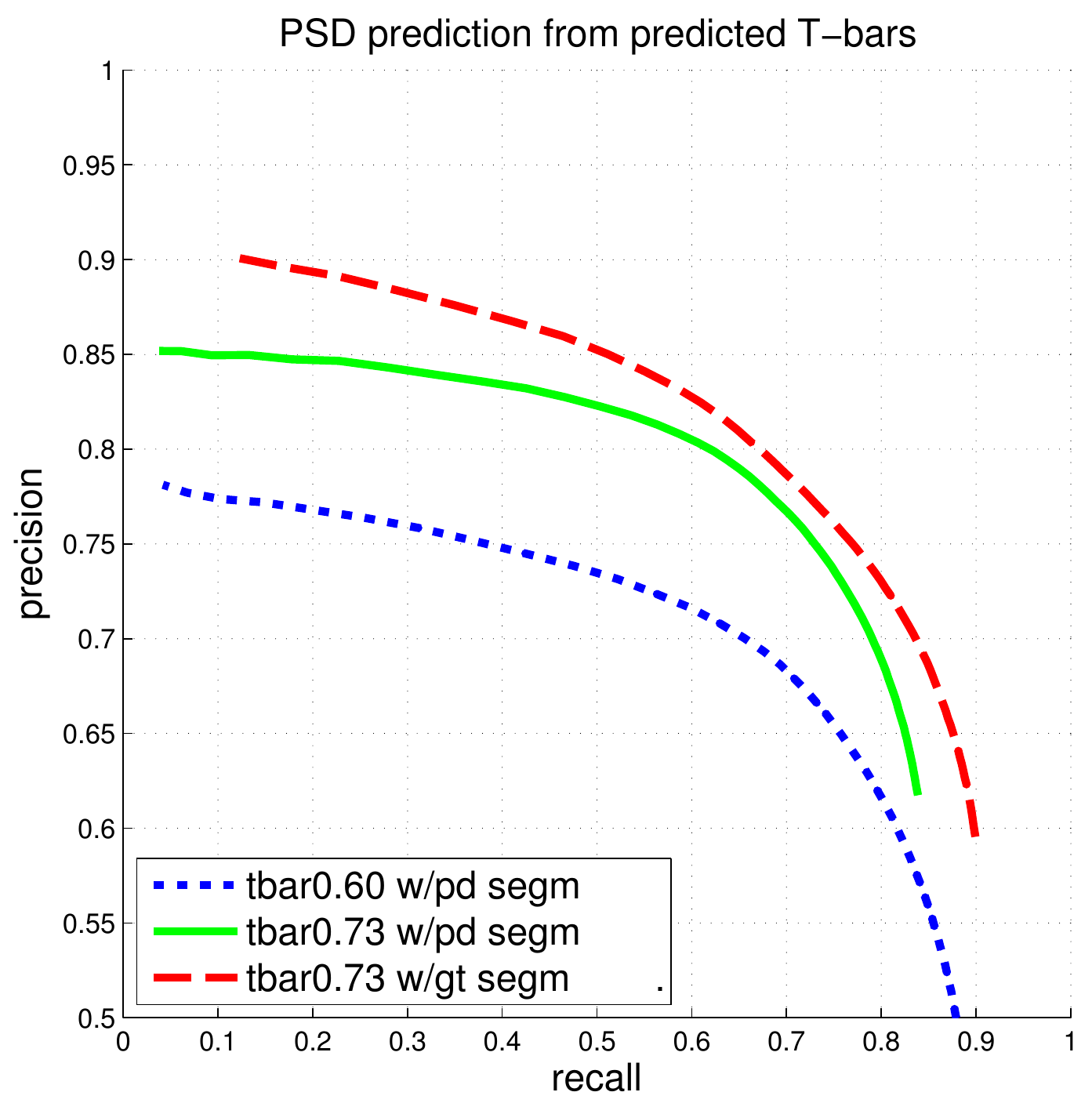}
\caption{\label{fig:fib25_psd} PSD precision/recall, where each PSD is considered separately (weighted view of connectome graph, see Section~\ref{sec:metrics}). Performance is computed both when access to the ground-truth segmentation is available during PSD detection (ground-truth/gt segm), and when only the fully-automated, predicted segmentation is available during PSD detection (predicted/pd segm). \textbf{Left:} Plot of PSD prediction performance in isolation, using ground-truth T-bar locations.  \textbf{Right:} Plot of end-to-end performance, using predicted T-bar locations: using a larger set of predicted T-bars with confidence greater than 0.60 (tbar0.60), and a smaller set of predicted T-bars with confidence greater than 0.73 (tbar0.73).}
\end{figure}

Next, we evaluate the performance of the PSD predictor.  We consider performance both under the scenario in which we have access to the ground-truth segmentation, and in which we only have access to a predicted, fully-automated segmentation.  We first make a note about the ``ground-truth segmentation''.  This segmentation was produced by starting from an automated segmentation (separate and distinct from the fully-automated segmentation we use for synapse prediction), and manually proofreading the segmentation by applying merge and split operations as necessary.  This ground-truth segmentation therefore aims to get the correct general topology, but is not refined to the point of necessarily assigning a correct label at the voxel level, and additionally may have orphan fragments that were not merged into larger bodies.  One important consequence is that, when we compute performance using this ground-truth segmentation, we typically ignore all predictions that fall into such orphan fragments, defined as segments that contain neither a ground-truth T-bar nor ground-truth PSD.  In other words, predictions that fall into such fragments are not counted when computing precision.  Additionally, we shift PSD point annotations using the same criteria as when shifting T-bar annotations as mentioned above.

We first evaluate PSD prediction assuming that we have access to ground-truth T-bar locations, in order to evaluate the performance of the PSD detector on its own.  This performance is given in the left plot of Figure~\ref{fig:fib25_psd}.  Next, we evaluate PSD prediction using predicted T-bar locations.  We consider two possible sets of T-bar predictions, one at a more conservative threshold of T-bars with confidence above 0.60, and a more aggressive threshold of T-bars with a confidence above 0.73.  For both cases, we compute precision/recall considering each PSD separately, corresponding to a weighted view of the connectome graph.  Importantly, we note that although performance is best when the ground-truth segmentation is available during PSD prediction, our PSD predictor is still able to achieve close performance using the automated, predicted segmentation.  

\subsection{End-to-end Synapse Performance and Comparison}

\begin{figure}
\centering
\includegraphics[width=0.45\textwidth]{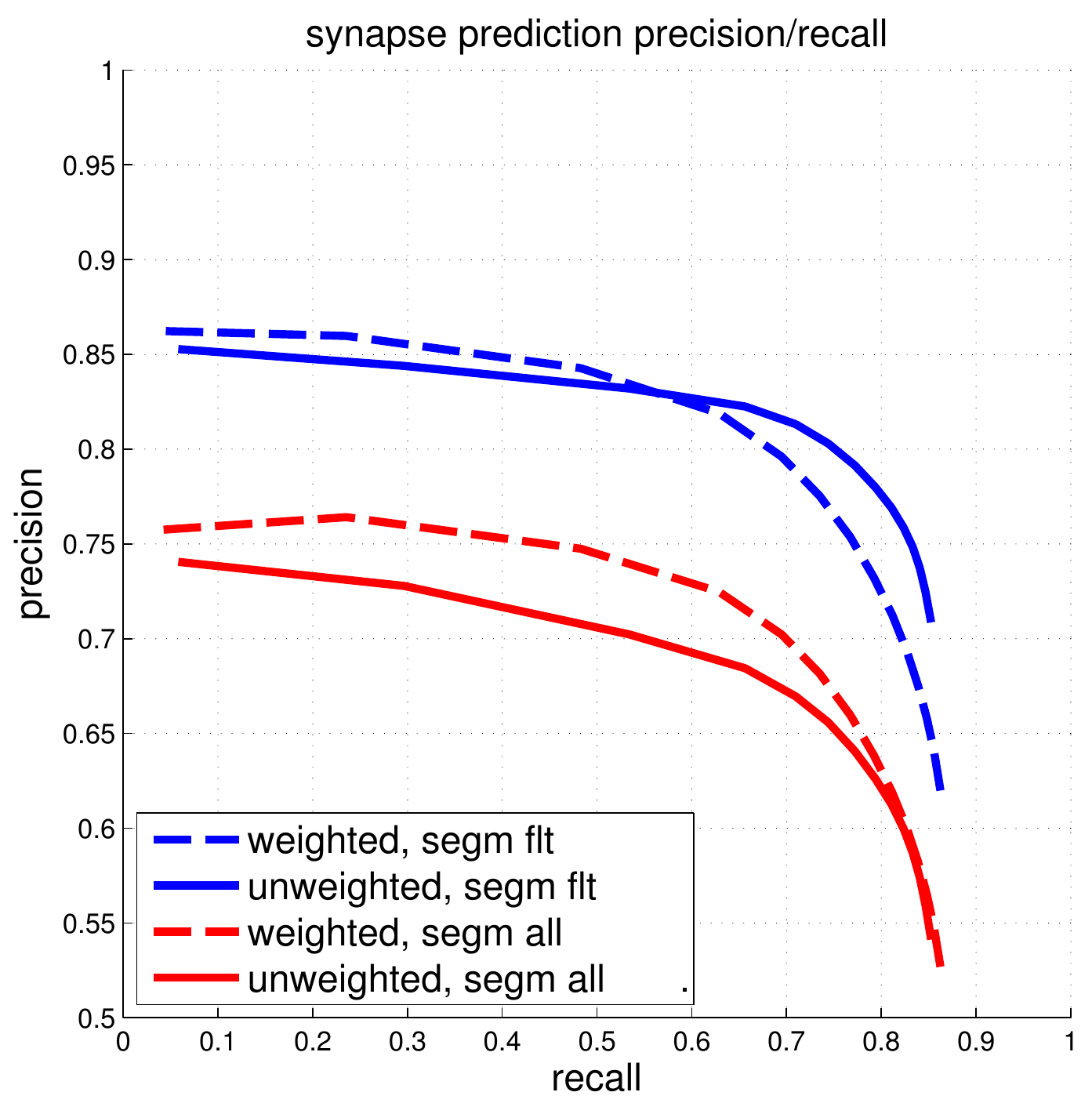}
\includegraphics[width=0.45\textwidth]{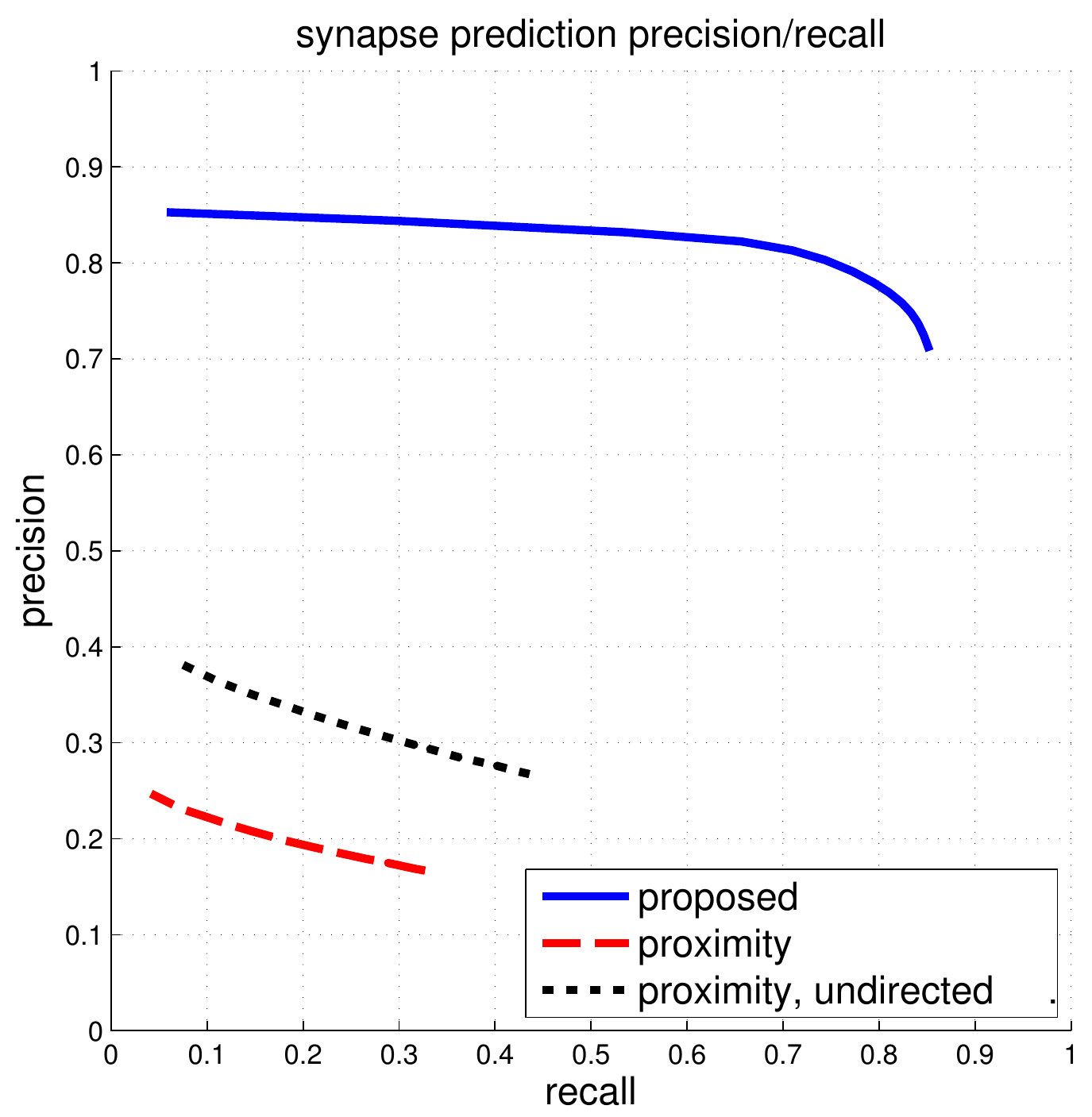}
\caption{\label{fig:fib25_syn} Global connectome graph precision/recall.  \textbf{Left:} The top blue curves show PR of the connectome graph, with the dashed curve computing PR using a weighted view of the graph edges, and the solid curve computing PR using a unweighted binary view of the graph edges.  These curves are computed using the filtered set of bodies in the ground-truth segmentation, as described in Section~\ref{sec:psd_perf}.  For reference, the bottom red curves show weighted and unweighted PR if all bodies (adding in orphan segments) are considered. \textbf{Right:} Comparison against the baseline of using ground-truth body-proximity as a proxy for synaptic contact.  All curves show unweighted PR on the filtered set of bodies.}
\end{figure}

We next give a full end-to-end evaluation of our synapse detection pipeline, with respect to the final generated connectome.  As determining an acceptable prediction accuracy is difficult without considering the particular connectomics application domain, we present a range of performance curves using our proposed error metrics.  Additionally, we compare against a baseline using neuronal-body proximity/contact as a proxy for synaptic contact.  For this baseline, we use the \emph{ground-truth} segmentation.  We randomly sample points at boundaries between ground-truth segments, and then randomly select the direction of the synapse (pre- and post-synaptic bodies).  For this proximity-based comparison, we also compute precision/recall using an undirected view of the connectome graphs, thereby allowing for matches even if the predicted direction of synapse was incorrect.  

The left plot of Figure~\ref{fig:fib25_syn} gives the precision/recall of our proposed system, using the fully-automated predicted segmentation.  We fix the threshold for T-bars at 0.73, accepting all T-bars above this threshold, and vary the threshold for the PSD detector to generate PR curves, under both a weighted and unweighted view of the connectome graph edges.  The right plot shows a comparison against the baseline using body-proximity as a proxy for synaptic contact.  Even using the ground-truth segmentation, and computing the undirected edge PR, this baseline performs much worse.

\begin{figure}
\centering
\includegraphics[width=0.45\textwidth]{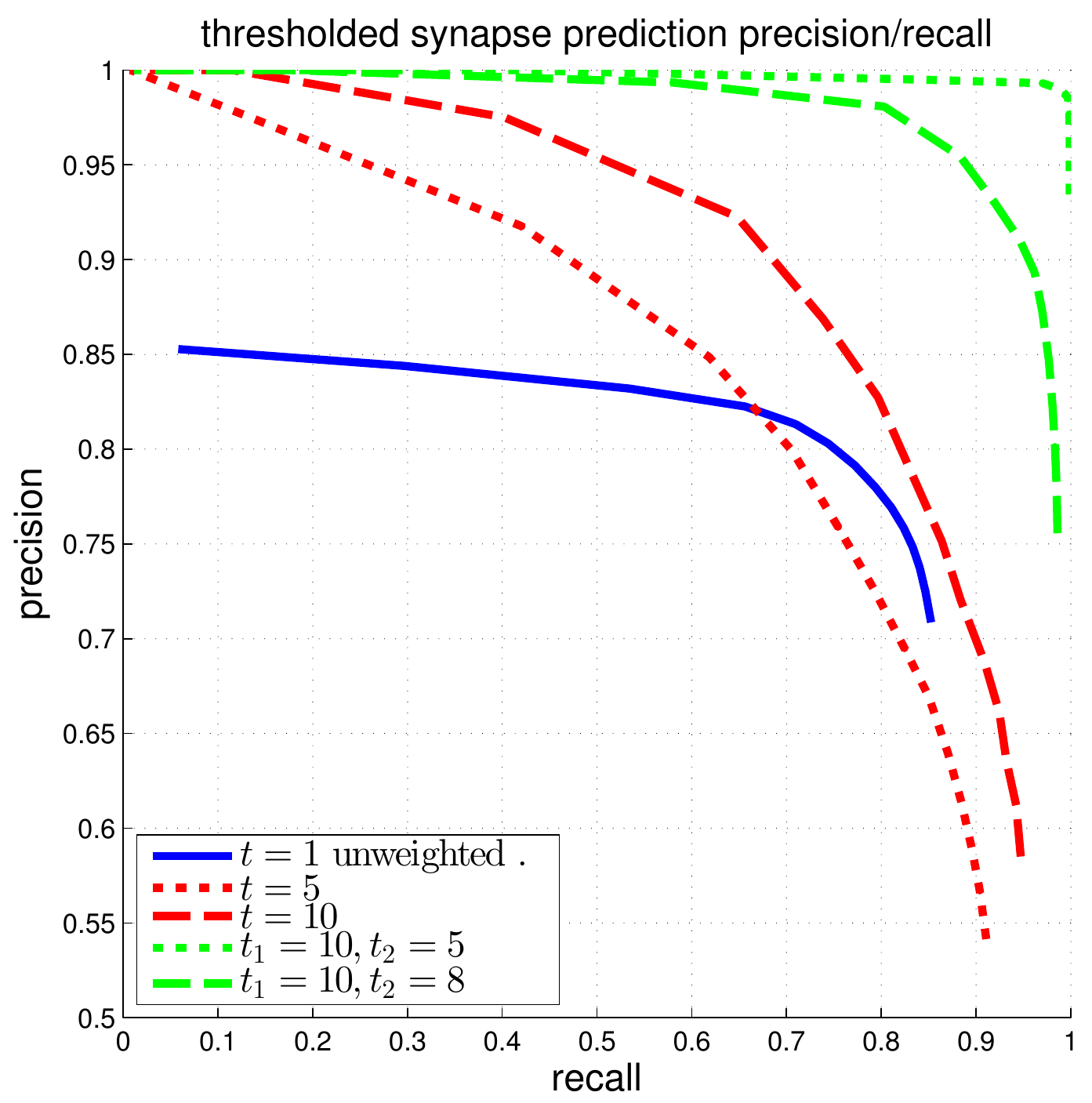}
\includegraphics[width=0.45\textwidth]{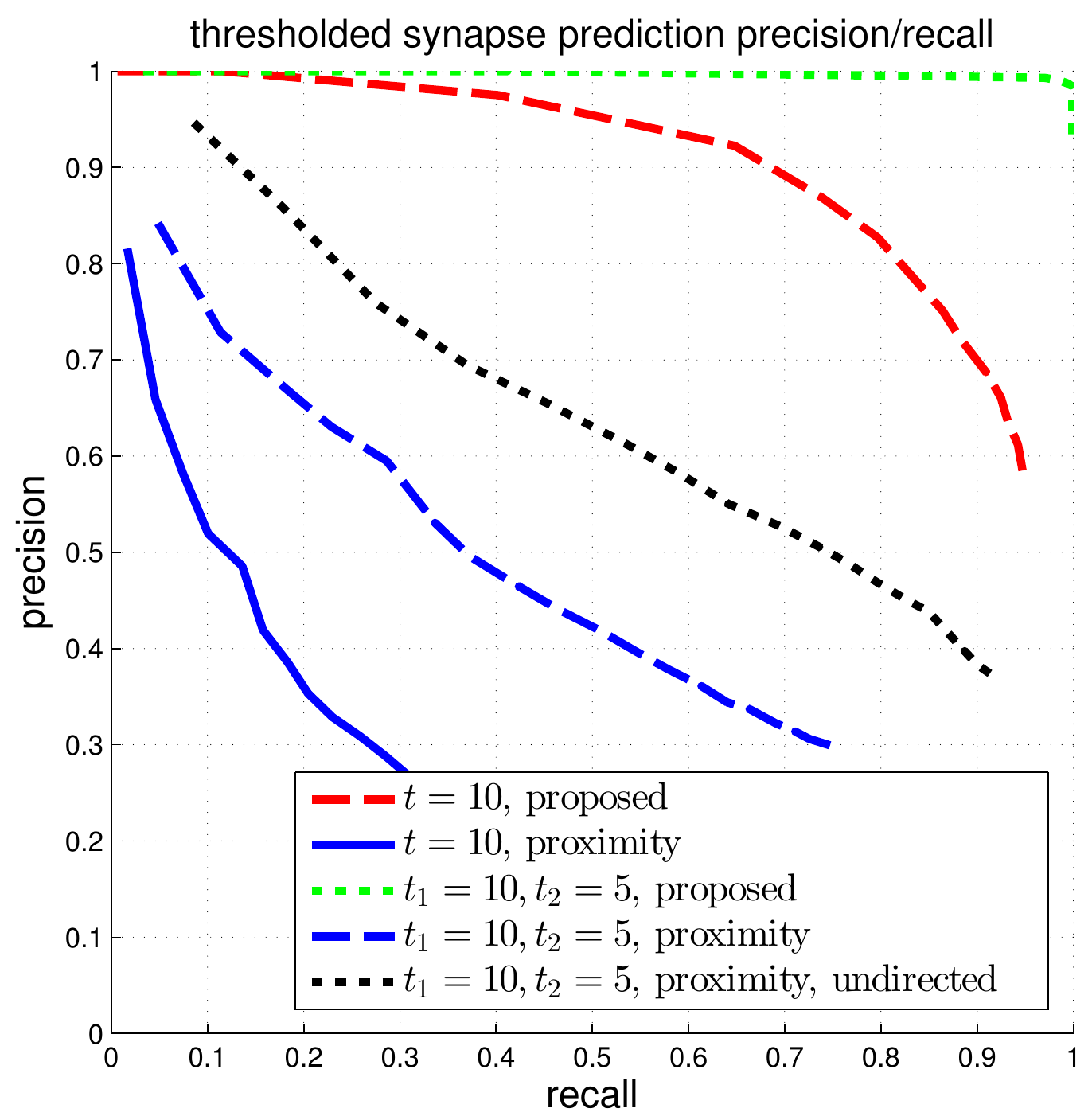}
\caption{\label{fig:fib25_syn_thresh} Thresholded global connectome graph precision/recall.  \textbf{Left:} The solid blue curve shows the unweighted PR from the previous figure, which is equivalent to a threshold of $t=1$.  The red curves give symmetric thresholded PR at $t=5,10$.  The green curves show asymmetric thresholded PR at $t_1=10,t_2=5,8$.  \textbf{Right:} Comparison against the baseline using body-proximity.  Focusing on strong error cases shows that while the proposed method only makes a few mistakes at $t_1=10,t_2=5$, the body-proximity baseline still performs comparatively poorly.}
\end{figure}

Next, we evaluate synapse detection performance using our proposed variants to precision/recall, as shown in Figure~\ref{fig:fib25_syn_thresh}.  We give curves when thresholding the edges at different values $t$, \ie, a (unweighted) edge is preserved in the connectome graph if the original edge weight is greater than $t$.  If $t=1$, then the curve is equivalent to the above unweighted graph PR.  We also give curves using our proposed asymmetric thresholded $t_1,t_2$ PR.  We again compare with the baseline of using body-proximity.  

\begin{figure}
\centering
\includegraphics[width=0.45\textwidth]{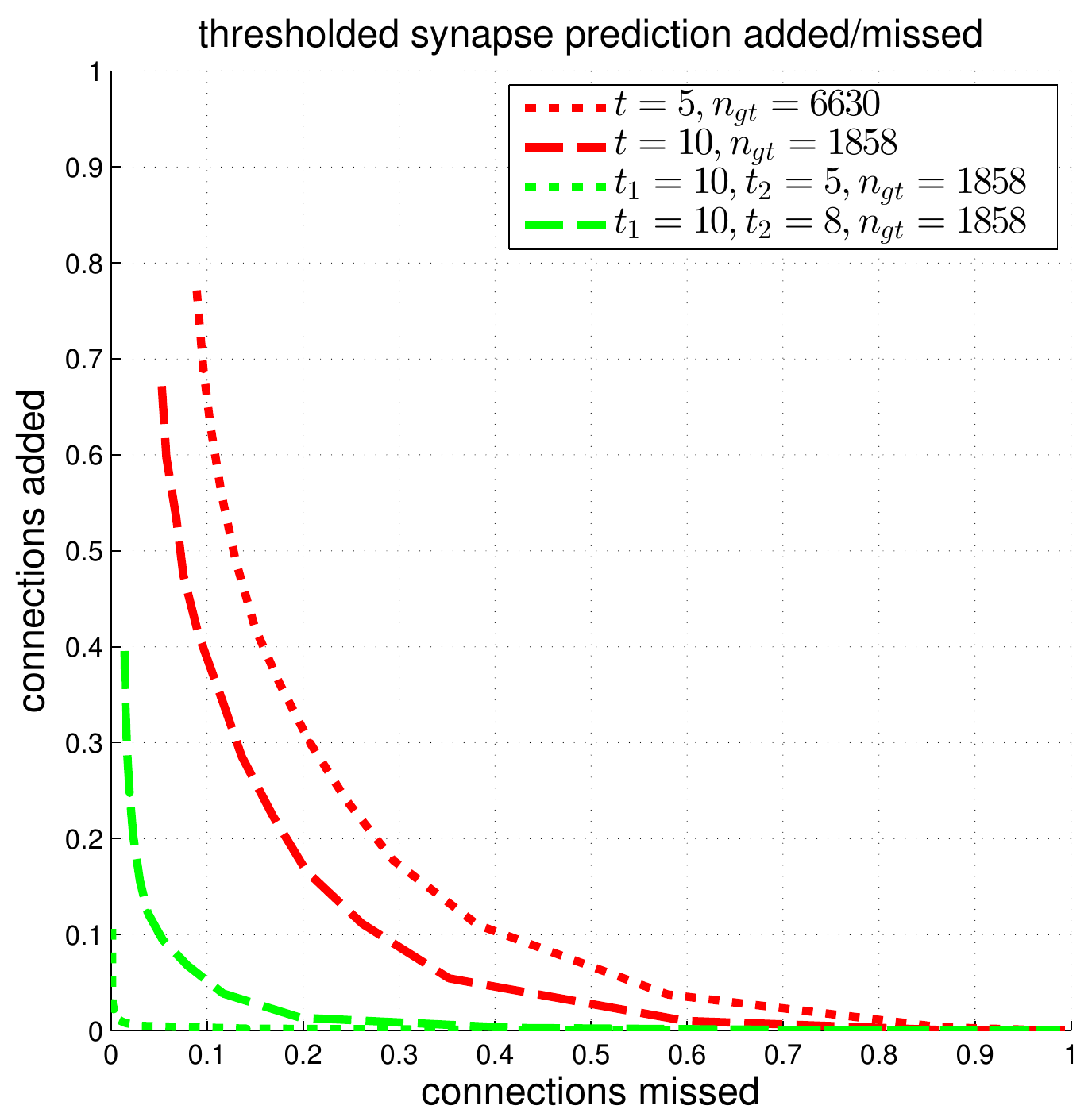}
\includegraphics[width=0.45\textwidth]{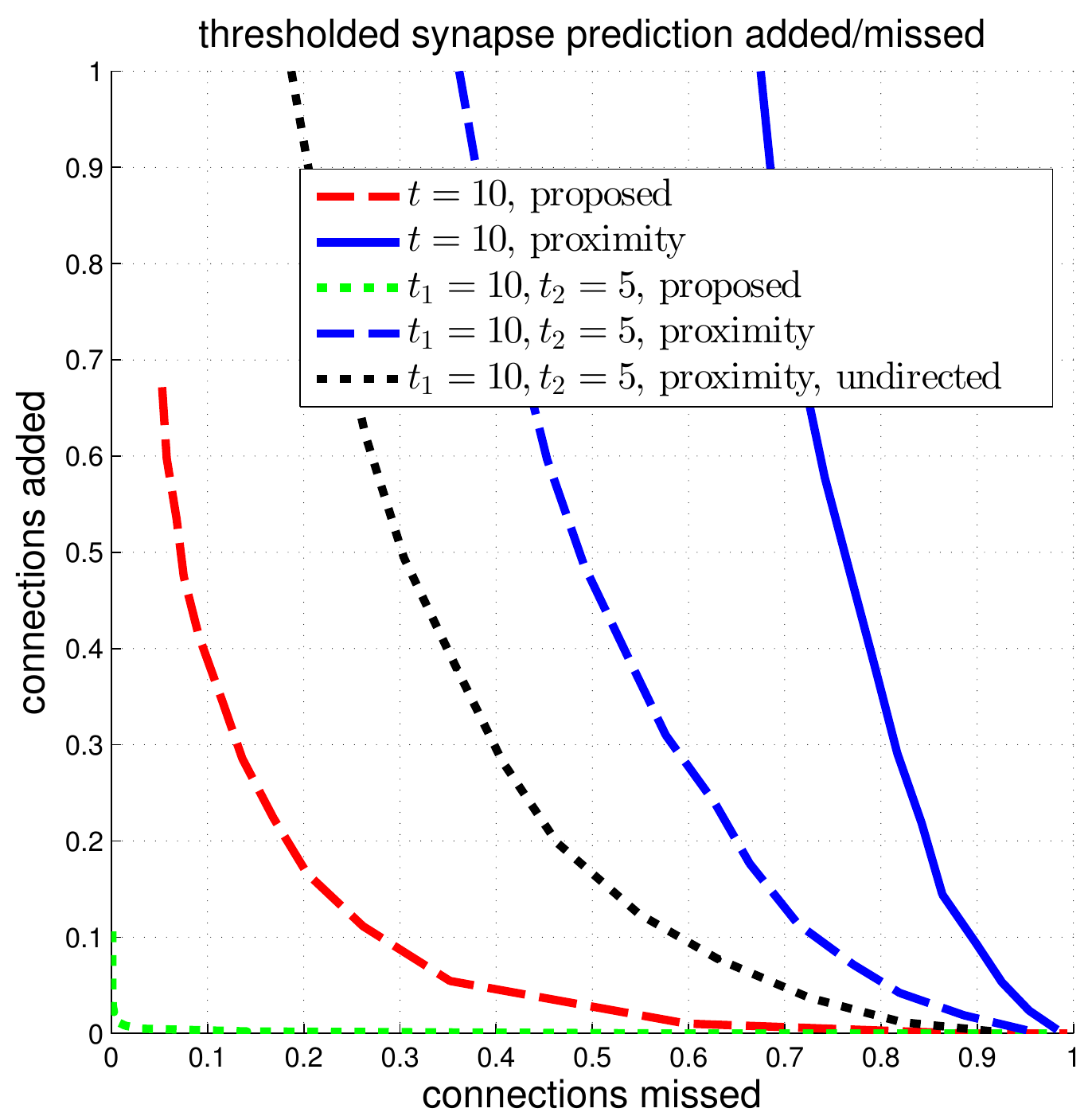}
\caption{\label{fig:fib25_syn_addmiss} Thresholded global connectome graph connections added/missed.  \textbf{Left:} The curves show errors in terms of connections that added and missed, using the same thresholds as the PR curves in Figure~\ref{fig:fib25_syn_thresh}.  \textbf{Right:} Comparison against the baseline using ground-truth body-proximity, both directed and undirected edges.}
\end{figure}

We give another view of synapse detection performance, using our metrics of connections strongly added and missed, in Figure~\ref{fig:fib25_syn_addmiss}.  For the case of thresholding with $t_1=10,t_2=5$, we have a total of about 2000 edges in the ground-truth connectome with a weight of at least $t_1=10$.  Using our proposed system, we can recover approximately 99\% of these edges (1\% connections missed) while introducing only 1\% falsely added connections.  By comparison, from the right plot in Figure~\ref{fig:fib25_syn_addmiss}, we can see that using body-proximity as a proxy for synaptic connection, when thresholding by $t_1=10,t_2=5$ and considering the directed graph, the normalized number of connections added and missed is approximately 50\%/50\%.  Therefore, even with this error metric that focuses on clearer, more ambiguous errors, this baseline approach is missing half the ground-truth connections and adding in approximately the same number of false connections.

\begin{figure}
\centering
\includegraphics[width=0.45\textwidth]{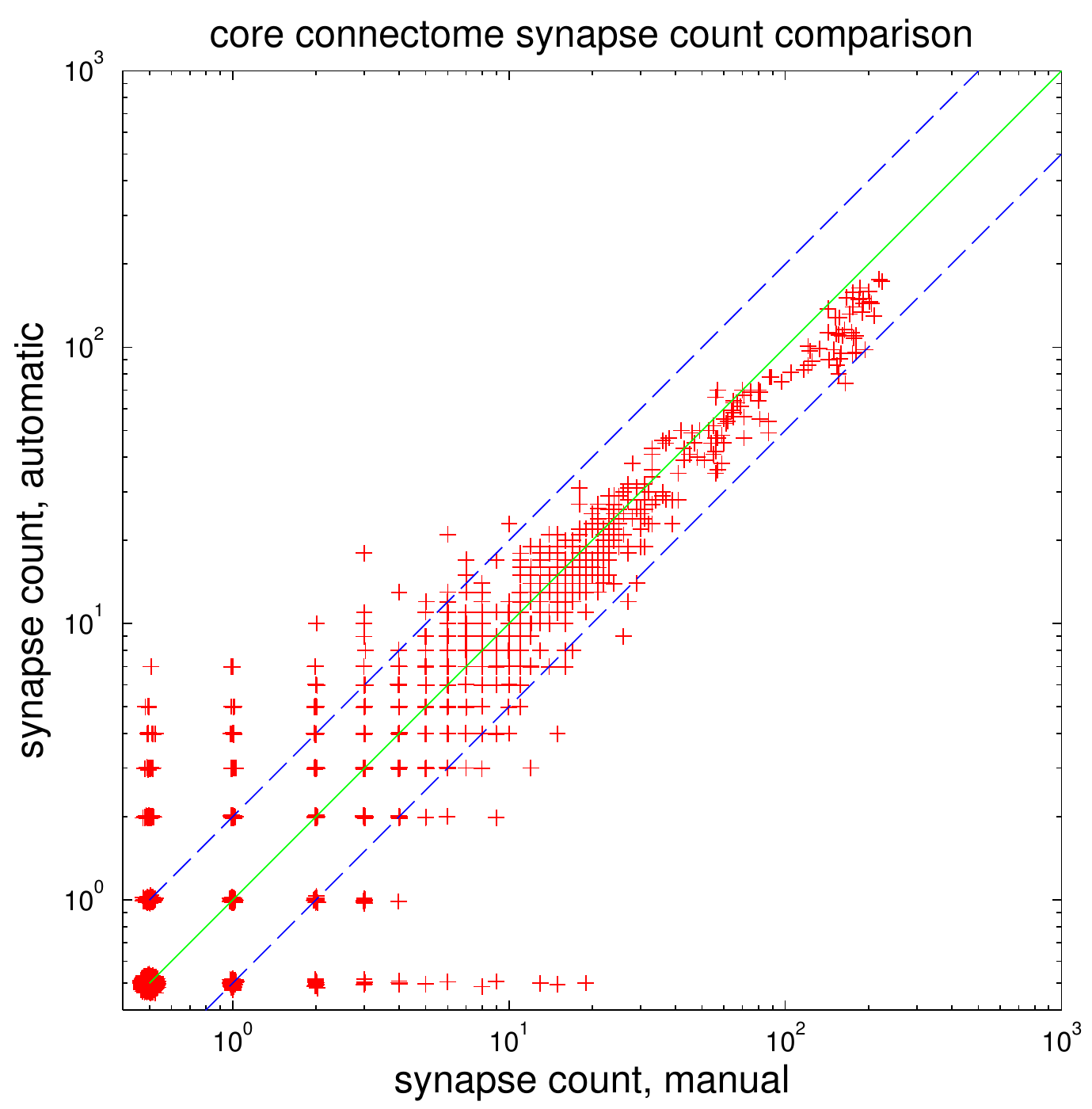}
\includegraphics[width=0.45\textwidth]{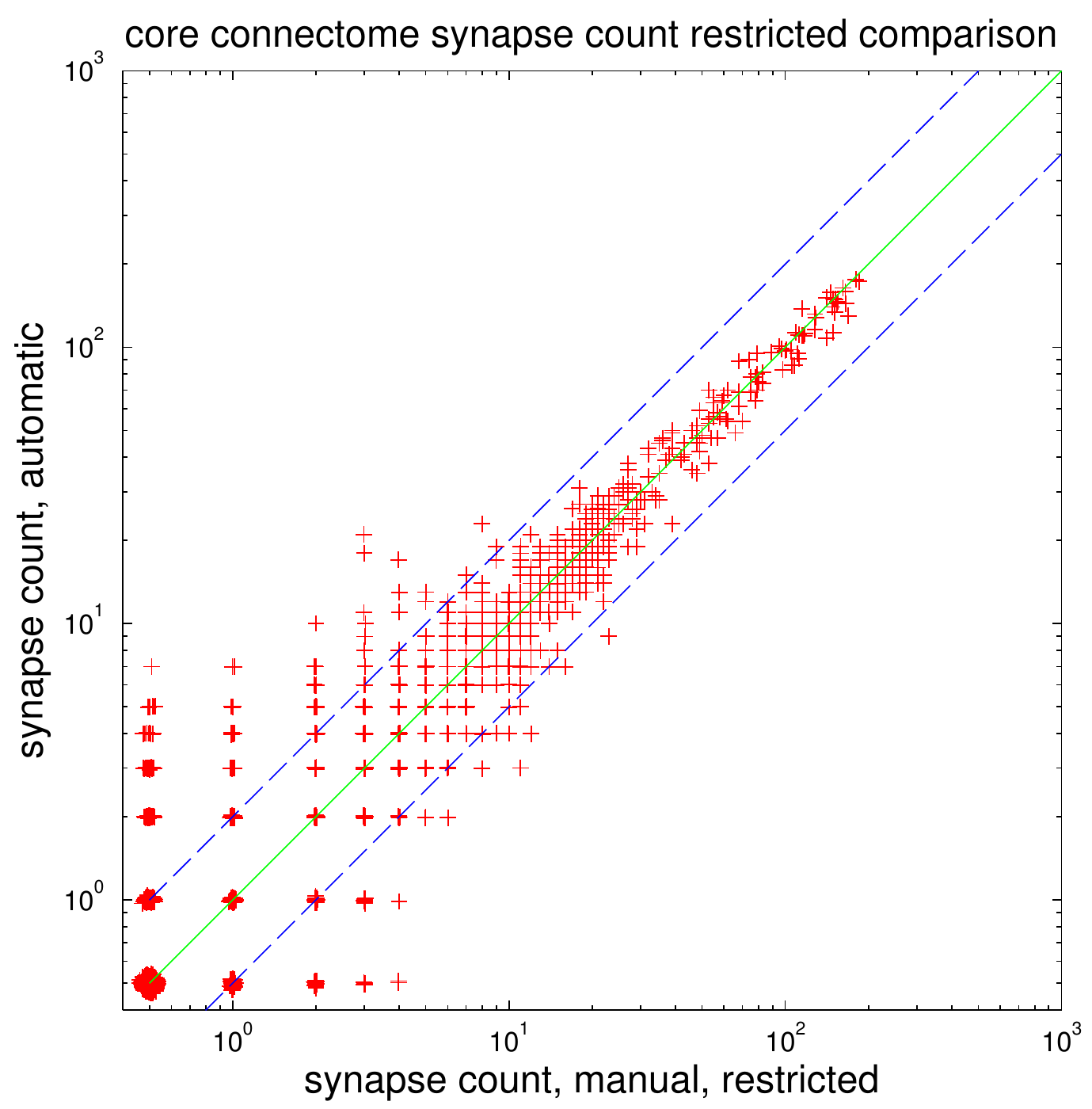}
\caption{\label{fig:fib25_count} Comparison of manual and automatic synapse counts, where each point indicates the counts for an edge in the connectome.  Edge weights of zero have been shifted to 0.5 to appear on the log plot.  Dashed lines indicate $y=2x$ and $y=\frac{1}{2}x$.  \textbf{Left:} Comparison with full ground-truth connectome.  \textbf{Right:} Comparison after removing ground-truth autapses and collapsing multiple connections from one T-bar to the same post-synaptic body.}
\end{figure}

Lastly, we present plots comparing automatic versus manual synapse counts when restricting edges to a core set of bodies and connectome, used in a study by Takemura~\etal~\cite{Takemura2015}.  Figure~\ref{fig:fib25_count} gives scatter plots, where each point gives the automated and manual synapse count for an edge in the connectome.  As mentioned above in Section~\ref{sec:psd}, our proposed system has limitations in that it does not attempt to predict autapses, and predicts at most one connection from a T-bar to a given post-synaptic body.  We therefore also give a comparison of automatic versus manual counts, shown in the right, after removing autapses and collapsing multiple connections from a single T-bar to the same post-synaptic body.  

We note that, for strong edges with a synapse count of 25 or above, our automated predictions fall within the indicated bounds of $y=2x$ and $y=\frac{1}{2}x$.  We can also examine edges in the automated and ground-truth connectome for which the corresponding connectome has a synapse count of zero.  We can thus see that, for all edges with a manual synapse count of at least 5, we are able to recover the edge, in the sense that the automated prediction gives a synapse count of at least one.  Conversely, for all edges with an automated synapse count of at least 8, the edge appears in the ground-truth connectome, as the manual synapse count is at least one.

\section{Conclusions}

In this paper, we have proposed an end-to-end system for automatic synapse detection in EM image data, capable of handling the polyadic synapses found in Drosophila.  We have additionally proposed a set of metrics to better assess the quality of a set of synapse predictions and whether such predictions are sufficiently accurate to be of use in connectomic studies.  We evaluate our system on the Drosophila seven column medulla data set, and show that it is capable of reconstructing high multiplicity synaptic connections, preserving biological pathways, while only making a small number of clear errors, and greatly outperforms a baseline using body proximity as a proxy for synaptic connection.  

One current limiting factor in our proposed system is the amount of run-time required for inference, as we scale analysis to larger image volumes.  We have experimented with producing synapse predictions for image volumes as large as several tera-voxels; in these case, distributed inference over a cluster of up to two thousand nodes took on the order of several weeks.  The majority of this time is spent on the initial T-bar detection step.  However, in on-going work, we have experimented with a convolutional architecture for T-bar detection, and traded off width (number of feature maps) for increased depth, and have seen improvements in speed while maintaining performance.\footnote{Code maintained and available at \url{https://github.com/janelia-flyem/flymatlib}.}

{\small{\bibliographystyle{ieee}
\bibliography{huangg.bib}
}}

\end{document}